\documentclass{article}

\usepackage{PRIMEarxiv}

\usepackage[utf8]{inputenc} 
\usepackage[T1]{fontenc}    
\usepackage{hyperref}       
\usepackage{url}            
\usepackage{booktabs}       
\usepackage{amsfonts}       
\usepackage{nicefrac}       
\usepackage{microtype}      
\usepackage{lipsum}
\usepackage{fancyhdr}       
\usepackage{graphicx}       
\graphicspath{{media/}}     
\usepackage[square, comma, sort&compress, numbers]{natbib}
\usepackage{multirow}
\usepackage{amsmath}
\usepackage{xcolor}

\definecolor{lightpink}{RGB}{255, 240, 245}
\definecolor{darkpink}{RGB}{219, 112, 147}
\definecolor{highlight}{RGB}{255, 69, 0}
\definecolor{titlebg}{RGB}{219, 80, 120}    
\definecolor{titlefg}{RGB}{255, 255, 255}   
\definecolor{grey}{RGB}{128, 128, 128}
\definecolor{mygreen}{RGB}{34, 139, 34}

\title{RLDBF: Enhancing LLMs Via Reinforcement Learning With DataBase FeedBack
}

\author{
  Weichen Dai,
  \And Zijie Dai,
  \And Zhijie Huang,
  \And Yixuan Pan,
  \And Xinhe Li, \\
  \AND Xi Li,
  \And Yi Zhou\thanks{The Corresponding Author, \url{yi_zhou@ustc.edu.cn}},
  \And Ji Qi,
  \And Wu Jiang
}

\begin{document}
\maketitle

\begin{abstract}
While current large language models (LLMs) demonstrate remarkable linguistic capabilities through training on massive unstructured text corpora, they remain inadequate in leveraging structured scientific data (e.g., chemical molecular properties in databases) that encapsulate centuries of accumulated scientific expertise. 
These structured datasets hold strategic significance for advancing AI for Science yet current approaches merely treat them as auxiliary supplements to unstructured text. 
This study pioneers a systematic investigation into enhancing LLMs with structured scientific data, using chemical molecular science as a testbed. 
We investigate the impact of incorporating molecular property data on LLM across distinct training phases, including continual pre-training, supervised fine-tuning, and reinforcement learning. 
Notably, to address the inherent limitation of numerical insensitivity in large models, we propose an innovative methodology termed "Reinforcement Learning with Database Feedback" (RLDBF). 
Experimental evaluations demonstrate the efficacy of the proposed approach, with the model exhibiting remarkable generalization capabilities on previously unseen data and other chemical tasks. 
The results substantiate the potential of our method in advancing the field of structured scientific data processing within LLMs.
\end{abstract}


\section{Introduction}
In recent years, artificial intelligence has demonstrated formidable capabilities across various tasks, achieving performance levels comparable to or even surpassing top human expertise - whether in recognition/generation tasks for speech~\citep{chu2023qwen} and images~\citep{Foret2020SharpnessAwareMF}, or knowledge-intensive tasks such as translation~\citep{Silver2017MasteringTG}, Go strategy~\citep{Takase2021LessonsOP}, and mathematical olympiad problem solving~\citep{Trinh2024SolvingOG}. 
The emergence of large language models (LLMs) has particularly propelled AI capabilities to unprecedented heights, catalyzing a new wave of artificial general intelligence (AGI) development.

The training paradigm for LLMs typically comprises three critical phases: 
pretraining, supervised fine-tuning (SFT), and reinforcement learning. 
Pretraining primarily enhances the model's knowledge, SFT empowers superior instruction-following capabilities, while reinforcement learning aligns model responses with human values through preference optimization.
Although LLMs exhibit remarkable knowledge capacity through pretraining on massive unstructured data, the inherent limitations of unstructured text - including low information density and ambiguous knowledge representation - hinder precise acquisition of domain-specific knowledge. 
Structured databases offer a promising solution to this challenge. 
Taking PubChem~\citep{kim2019pubchem}, a chemical database as an example, it contains over 100 million molecular entities, each characterized by standardized fields (e.g., SMILES~\citep{weininger1988smiles}, physicochemical properties, reaction pathways) that form an intricate knowledge network. 
This structured representation inherently achieves significantly higher information density compared to unstructured text.

Structured databases often encapsulate meticulously curated knowledge and experimental data of immense potential value - painstakingly accumulated by generations of top scientists. 
These resources contain substantial under-explored information. 
However, existing methodologies fail to fully exploit this wealth of data. 
Representative works like MoLFormer~\citep{ross2022large} primarily utilize molecular representations for pretraining while neglecting more critical knowledge dimensions such as molecular properties.
Meanwhile, reinforcement learning, though less data-hungry than pretraining and SFT, remains computationally intensive due to the substantial human annotation required for its preference data.

To address these dual challenges, we propose RLDBF (Reinforcement Learning with Database Feedback), a novel framework that synergizes structured databases with reinforcement learning. 
Our principal contributions are threefold:
\begin{itemize}
    \item We pioneer the concept of leveraging database data to alleviate LLMs' data bottleneck issues, providing new perspectives for exploiting knowledge in structured databases. 
    This may inspire future exploration of database-driven knowledge enhancement.
    \item We propose the first reinforcement learning framework integrated with structured databases for LLM enhancement. 
    Our meticulously designed architecture enables efficient knowledge absorption from databases while substantially reducing annotation overhead.
    \item Multi-dimensional experiments demonstrate our method's effectiveness. 
    Notably, models trained with RLDBF not only exhibit superior generalization on chemical tasks but also reveal emergent capabilities for assisting in-depth chemical reasoning.
\end{itemize}

\section{Related Works}

\subsection{AI for Chemistry}

The chemistry and materials science communities have witnessed the growing adoption of Transformer-based language models for various downstream tasks, including crystal structure prediction and coulombic efficiency prediction. These approaches can be broadly categorized by their sequence processing strategies: 1) post-training on foundation large language models (LLMs) with data encoded via pre-trained tokenizer, or 2) developing task-specific tokenizer and encoding schemes coupled with custom-defined models.

As demonstrated by \citet{dagan2024getting}, the distributional mismatch between natural language-trained tokenizer and domain-specific tasks can significantly degrade downstream performance. This has driven researchers to design specialized sub-word vocabularies and train compact Transformer-based architectures for specific applications. For instance, \citet{antunes2024crystal} developed a 25M-parameter GPT-like model trained on crystallographic information for structure generation. \citet{liu2025alchembert} achieved superior materials property prediction through a 110M-parameter BERT model trained on crystal structures and natural language descriptions. Three-dimensional generative capabilities were demonstrated by \citet{flam2023language}, while \citet{flam2023atom} pioneered protein-drug conjugates generation using specialized architectures.

However, the prohibitive computational resource requirements for large-scale pre-training have constrained model development. Post-training paradigms offer more accessible alternatives, benefiting from scaling laws where larger models capture more information. \citet{gruver2024fine} leverages post-trained LLMs to comprehend and validate three-dimensional atomic configurations, demonstrating the scaling law governing model capacity and structural fidelity in preserving critical crystal symmetries.

Recent efforts have focused on domain adaptation through curated datasets (e.g., textbooks, research papers, and QA pairs) to create chemistry-specific LLMs. \citet{zhang2024chemllm} fine-tuned \citet{cai2024internlm2} with 7M proprietary chemical instructions for chemical QA tasks. \citet{zhao2024chemdfm} conducted incremental pretraining and instruction tuning on Llama2 13B using extensive chemical corpora. Specialized applications include GPT-3 fine-tuning for electrochemical property prediction by \citet{xie2023large}, and text-encoded atomistic data adaptation of Llama2 7B-70B for stable crystal generation by \citet{gruver2024fine}, demonstrating that model scales positively correlated with symmetry preservation capabilities.

\subsection{Structured Data Integration}

Integrating structured/sub-structured data into natural language pre-trained LLMs presents unique challenges. Conventional approaches often naively serialize structured data into text sequences, as exemplified by \citet{ren2024watergpt}. However, \citet{mirchandani2023large} revealed through the ARC benchmark\citep{chollet2019measure} that LLMs exhibit varying sensitivity to structural semantics, complicating direct alignment between structured data representations and textual embeddings.

To bridge this semantic gap, \citet{sun2021ernie} incorporated knowledge graph triples via autoencoder networks, though scalability remains constrained by structured data requirements. Alternative strategies leverage LLMs to generate synthetic text from structured input, as demonstrated in the synthesis of clinical data by \citet{kweon2023publicly}. Recent advances address hallucination in synthetic data through structured data verification, as proposed by \citet{chenknowledge}, enabling higher-fidelity QA pair generation.

\subsection{Numerical Domain Knowledge}

The integration of experimentally derived or computationally generated structured data, often requiring precision to multiple decimal places, exposes critical limitations in LLM numerical reasoning. \citet{zheng2025large} systematically identified deficiencies in chemical numerical tasks. This highlights the fundamental challenge of achieving high-precision numerical memorization and reasoning in AI4Science applications.

Emerging research quantifies these limitations: \citet{yuan2023well} documented systematic underperformance in numerical operations, while \citet{sakamoto2025development} revealed deficiencies in numerical common sense reasoning. Domain-specific challenges are addressed by \citet{mishra2020towards}, identifying chemically grounded numerical tasks (e.g., stoichiometric balancing, molecular weight calculation) where current models struggle without explicit integration of chemical knowledge.

\section{Methods}

\begin{figure}[t]
\begin{center}
\includegraphics[width=\textwidth]{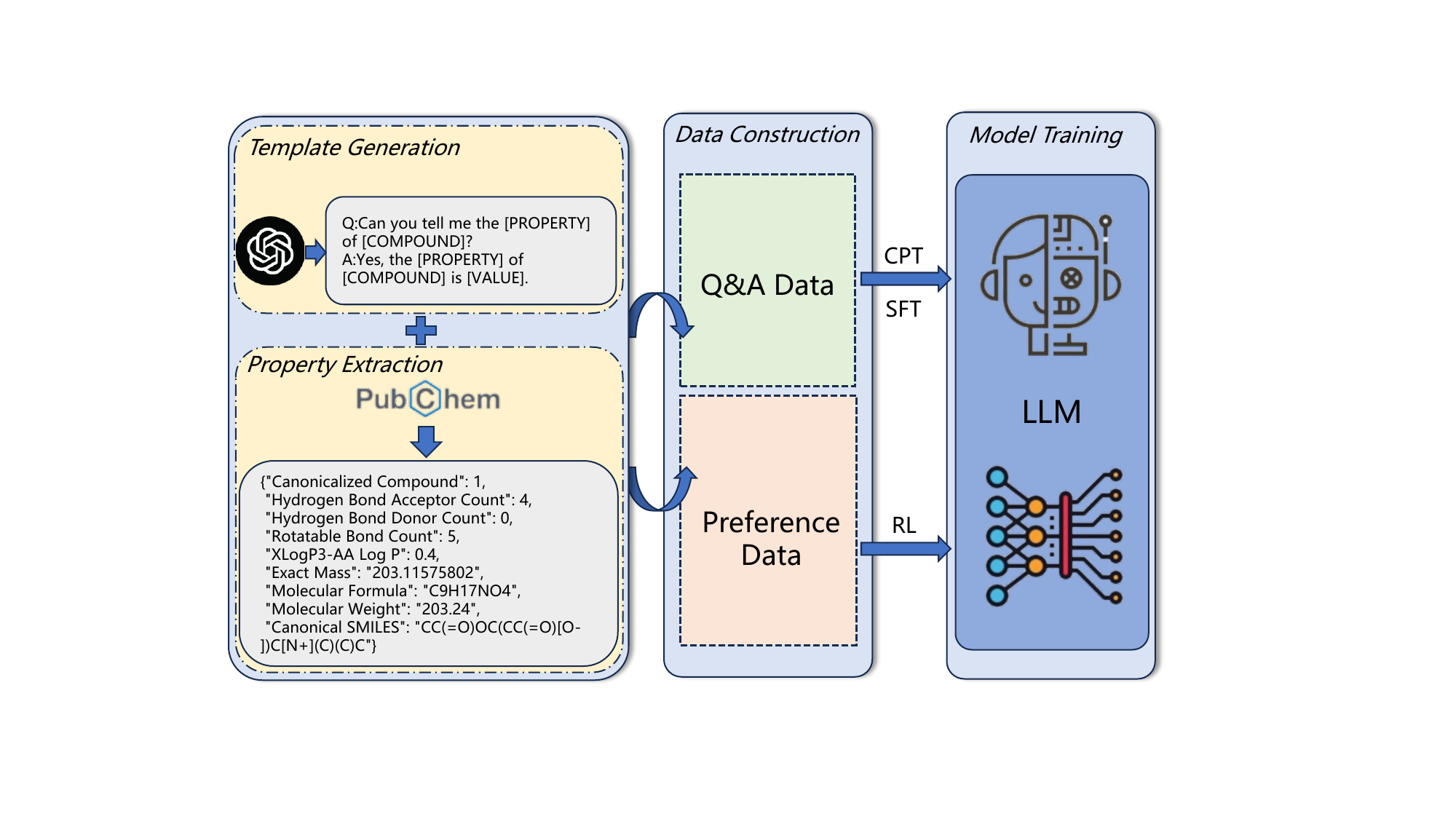}
\end{center}
\caption{Overview of the working pipeline.}
\label{framework}
\end{figure}

We demonstrate our pipeline for CPT, SFT and RL in Figure~\ref{framework}, including template genration, data construction and model training.

\subsection{Q\&A Template Construction}

The data in databases is highly structured, while LLMs are trained on unstructured text. 
A natural approach is to transform structured data into unstructured training data via template filling. 
We used GPT-4o to generate multiple Q\&A templates. 
To boost model performance on diverse instructions, GPT-4o was instructed to create templates beneficial for subsequent fine - tuning, ensuring diversity in style, structure, and expression. 
Next, we applied a three - step filter: 
1) For templates with identical questions, only one was kept; 
2) Templates with general yes - no questions were removed to prevent the model from only learning yes/no answers; 
3)Templates were manually checked, and those with unreasonable content or wording errors were deleted.
Finally, 219 templates were generated, examples are shown in appendix Table~\ref{gpt_templates}.
To increase the difficulty of subsequent tests, we split these templates into training and test sets. 
Specifically, we first used 
sentence-transformers\footnote{https://huggingface.co/sentence-transformers/all-MiniLM-L6-v2}~\citep{reimers-2019-sentence-bert} 
to embed each question template, then clustered them using the DBSCAN algorithm. 
We adjusted the algorithm's parameters to get a reasonable number of outliers, using the corresponding templates to construct the test set. 
Ultimately, we obtained 184 training templates and 35 test templates.

\subsection{CPT \& SFT Data Synthesis}

Continual Pre-Training (CPT) and SFT are two common methods for training LLMs. 
To explore the impact of different training data synthesis methods and establish a comprehensive benchmark for subsequent tests, we designed various ways to synthesize training data based on the templates from the previous subsection.
\paragraph{CPT Data}
The main purpose of CPT is to equip LLMs with more knowledge. 
We tried five methods to help LLMs better establish connections among compounds, properties, and values:
1) Directly concatenating Q\&A pairs;
2) Directly concatenating molecule, property, and value;
3) \& 4) Using question templates with numerical answers only. 
To help LLMs better connect questions with real answers and avoid them only outputting answers, we added "Answer the question only with the corresponding value" after each question.
This was also applied in synthesizing SFT and RL data. 
We then decorated Q\&A pairs with "User-Assistant" or "**";
5) Directly building data with "The [property] of [compound] is [value]" template.
\paragraph{SFT Data} 
We tried three methods:
1) Directly using Q\&A pairs;
2) Using Q\&A pairs with answer templates added to questions as prompts to reduce the impact on the model's instruction - following ability;
3) Using question templates with only numerical answers.
The synthesis methods above and examples are summarized in appendix Figure~\ref{cptsft_syn}.

\subsection{RLDBF}

\begin{figure}[t]
\begin{center}
\includegraphics[width=\textwidth]{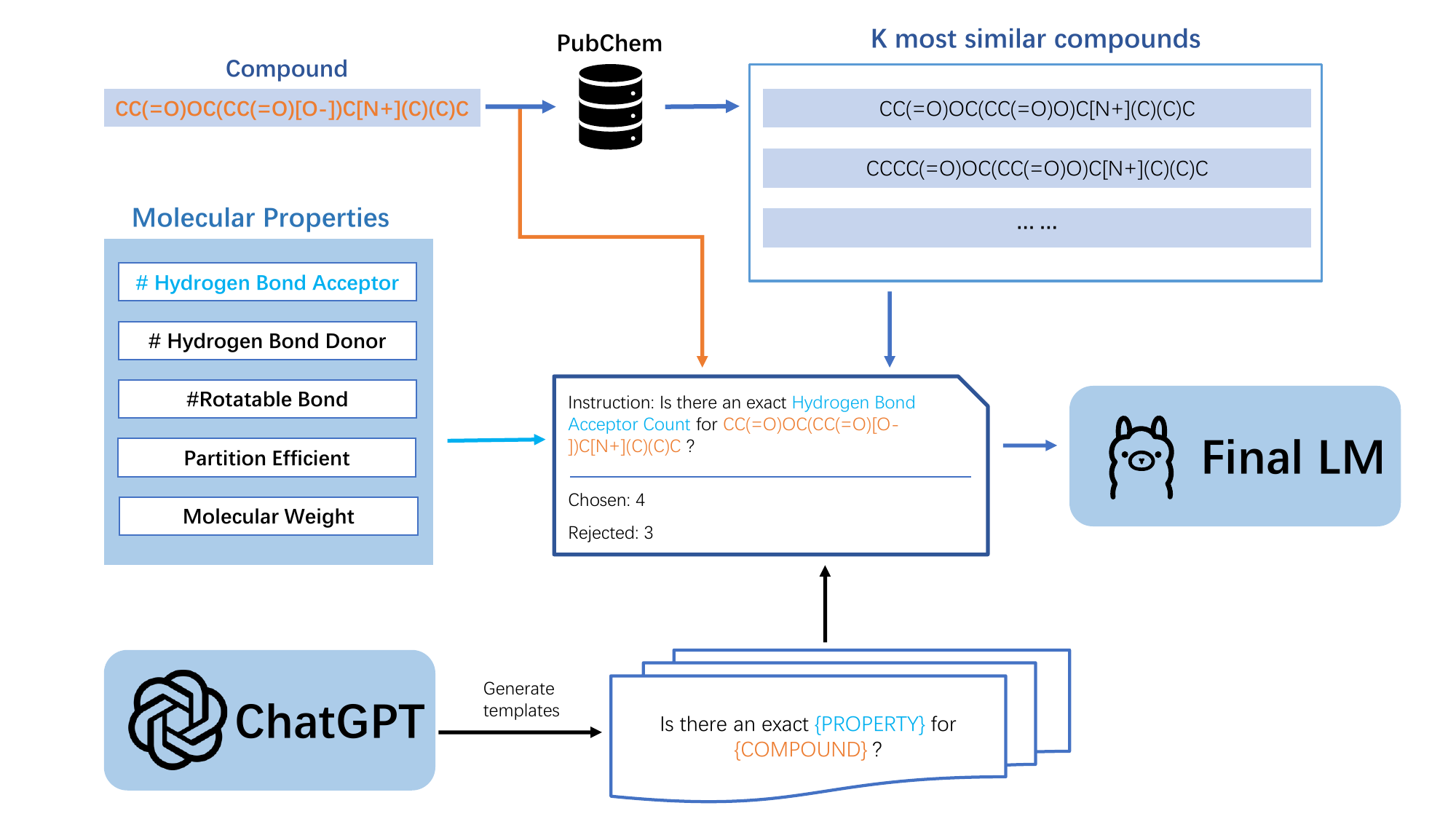}
\end{center}
\caption{Overview of RLDBF.}
\label{dpo_framework}
\end{figure}

\paragraph{Motivation}
The feedback in Supervised Fine-Tuning (SFT) is one-way, only indicating correct answers, not wrong ones. 
Though effective, it has limitations. 
Reinforcement Learning from Human Feedback (RLHF)~\citep{ouyang2022training} enables models to reject incorrect outputs, enhancing the safety and stability of their outputs. 
It allows models to learn both correct answers and how to avoid wrong ones.
Inspired by this, we decided to employ reinforcement learning to better leverage database data for model training. 
However, RLHF relies heavily on high quality human preference labels, which increases data acquisition costs and exposes model performance to annotator bias due to the lack of factual basis in data comparison. 
Considering this, we proposed Reinforcement Learning from DataBase Feedback (RLDBF), using databases to automatically build preference data.
Databases are natural data labelers. 
We can create many preference data pairs from different database instances and establish their priority by disturbing attributes and values.
Take these two data entries as an example: 
"[Ins. A]-[Attr. 1]-[Attr. 2]; [Ins. B]-[Attr. 1]-[Attr. 2]". 
Based on the severity of knowledge-based errors, the preference data can be constructed as follows:
\begin{itemize}
    \item Score 3 (result match): [Attr. 1] of [Ins. A] is [Attr. 1 Value of A]
    \item Score 2 (instance match, attribute mismatch): [Attr. 1] of [Ins A] is Value(Attr. 2, A)
    \item Score 1 (instance mismatch, attribute match): [Attr. 1] of [Ins A] is Value(Attr. 1, B)
    \item Score 0 (all mismatch): [Attr. 1] of [Ins A] is Value(Attr. 2, B)
\end{itemize}

\paragraph{Implementation}
RLAIF~\citep{lee2023rlaif} proposes to directly use the feedback from a LLM as the reward signal in RL.
This approach not only reduces the overhead associated with AI preference labeling and reward model training (which is essentially a form of distillation) but also demonstrates superior performance compared to learning reward models through distillation.
This finding aligns with our perspective, as training a reward model on database data may introduce additional biases.
To address this, we adopt a more streamlined approach through Direct Preference Optimization (DPO)~\citep{rafailov2023direct}, enabling direct discrimination between correct and erroneous data without intermediary models.
We enhance output fidelity by exclusively constructing preference pairs using verified numerical ground truths—even for "reject" samples—thereby minimizing noise from stochastic perturbations. 
Empirical analysis reveals that LLMs exhibit limited discriminative capacity for molecular structures due to multi-token compositional representations, frequently conflating properties of structurally analogous molecules. 
To mitigate this, we strategically generate preference pairs by contrasting target molecules' verified property values with those of their nearest structural analogs.
Specifically, for each specific molecule $C$, we first identify $K$ most similar molecule $\{S_i\}$.
Then, for a given property $P$ of molecule $C$, we construct $K$ preference pairs, where the "chosen" value in each pair is the ground truth value, and the "reject" value is the value of property $P$ from $\{S_i\}$. 
Finally, to minimize the influence of textual descriptors, the preference pairs contain only the specific numerical values without additional words.
These constructed pairs finally serve as the DPO data for model learning. 
Through the above steps, we have implemented the entire process of RLDBF.

\paragraph{Comparable Alternatives}
We also proposed several alternative strategies for constructing "reject" data, including:
1) Using values of different properties for the same molecule, values of the same property for different molecules, and values of different properties for different molecules (SMILES remains unchanged);
2) Using the same molecule with different properties, different molecules with the same property, and different molecules with different properties;
3) Only using the same molecule with different properties and different molecules with the same property;
4) Only using the same molecule with different properties and the most similar different molecule with the same property;
5) Adding answer templates to the instruction part based on 4);
6) Refining 5) by only using the most similar different molecule with the same property.
Examples are shown in appendix Figure~\ref{dpo_syn}.
These construction methods are also used as benchmarks for subsequent tests.

\section{Experiments}

\subsection{Dataset}

\subsubsection{Pubchem}

PubChem~\citep{kim2019pubchem} is an open-access chemical database, systematically integrates structural, physicochemical, and bioactivity data for small molecules from 750+ authoritative sources. 
Its triad architecture (Compound, Substance, BioAssay) enables cross-domain exploration of structure-biological activity relationships through curated experimental results and predictive models. 
PubChem has become fundamental infrastructure for computational chemistry and translational studies, ensuring reproducibility while bridging experimental and computational research paradigms.

In this study, we constructed datasets using downloadable PubChem data~\footnote{https://pubchem.ncbi.nlm.nih.gov/docs/downloads} 
due to its substantial volume and accessibility.
The processed data was partitioned into multiple files, each containing 500,000 molecules. 
We first examined the key-value pairs of each entry, selecting non-empty keys to ensure data integrity for subsequent experiments. 
Five molecular properties were ultimately chosen for dataset construction: 
Hydrogen Bond Acceptor Count, Hydrogen Bond Donor Count, Rotatable Bond Count, Octanol-water Partition Coefficient, and Molecular Weight. 
These properties can also be computationally derived. 
Additionally, SMILES~\citep{weininger1988smiles} was adopted as the molecular textual representation.

\paragraph{Training Set}
Under computational resource constraints, we maximized the training dataset size within feasible limits, ultimately constructing it based on the first 10,000 molecules from PubChem.

\paragraph{Test Set}
The test set was systematically designed along two dimensions with four difficulty levels. 
The two dimensions distinguish between in-domain molecules (from the training set) and out-domain molecules (from PubChem molecules beyond the 500,000th entry), enabling evaluation of both model learning capacity and generalization ability.
Each question requires the model to predict specific molecular properties, with four progressively challenging levels:
\begin{itemize}
    \item Level 1: Distractors involve basic numerical perturbations of the correct answer: $\pm0.1, \times10, \div10$
    \item Level 2: Sophisticated perturbations include nine variants: $\pm0.1, \pm1, \times(-1)$, and combined operations ($\times(-1)$ followed by $\pm0.1 / \pm1$)
    \item Level 3: Distractors comprise values from different properties of the same molecule, same properties of different molecules, or different properties of different molecules;
    \item Level 4: Distractors are selected from the most similar molecules' corresponding properties
\end{itemize}
To mitigate random bias, we implemented three independent sampling repetitions for each dimension-level combination, generating 200 questions per sampling instance. 
We also calculate the weighted sum for comparison:
\begin{equation}
    W.S. = 0.4*S_4 + 0.3*S_3 + 0.2*S_2 + 0.1*S_1,
\end{equation}
where $S_i$ is the sum of the average scores of Level i.

\subsubsection{ChemBench}

ChemBench~\citep{zhang2024chemllm} is a comprehensive benchmark designed to evaluate the performance of AI models in chemistry-related tasks. 
It encompasses a diverse set of problems, including 
Name Conversion(NC), Property Prediction(PP1), Mol2caption(M2C), Caption2mol(C2M), Product Prediction(PP2), Retrosynthesis(RS), Yield Prediction(YP), Temperature Prediction(TP) and Solvent Prediction(SP).
This benchmark provides a rigorous assessment of model capabilities in the chemical domain, and facilitates standardized comparisons across different approaches, promoting advancements in AI-driven chemistry research.
Results are rounded to the nearest tenth.

\subsection{Training Setup}
Our experiments were conducted on a computing cluster equipped with 8×NVIDIA A100 80GB GPUs, utilizing the Llama3-8B-Instruct~\citep{grattafiori2024llama} model as the base model. 
To optimize experimental efficiency, we initially performed preliminary investigations to determine appropriate learning rate ranges. 
The final learning rates for both the CPT and RL were selected from [1e-7, 5e-7, 1e-6], while SFT employed learning rates from [1e-7, 1e-6]. 
For each candidate learning rate configuration, we trained corresponding model instances for 1 epoch and subsequently selected the optimal performing model for comparative analysis. 
Throughout the training process, all model parameters remained activated and trainable, the DeepSpeed ZeRO-1\citep{rasley2020deepspeed} are employed.

\subsection{Baselines}
In addition to the trained models discussed in previous sections, we conducted comprehensive comparative analyses with other prominent LLMs within our benchmarking framework. 
These models include: 
GPT-4o-mini~\citep{hurst2024gpt},
GPT-3.5-Turbo,
Qwen2-7B-Instruct~\citep{yang2024qwen2technicalreport},
Llama3-8B-Instruct; 
and domain-specific chemical models comprising
KALE-LM-Chem~\citep{dai2024kale}, 
ChemDFM~\citep{zhao2024chemdfm},
ChemLLM~\citep{zhang2024chemllm},
and LlaSMol-Mistral-7B~\citep{yu2024llasmol}.

\subsection{Results}

\begin{table}[t]
\begin{center}
\begin{tabular}{l|c|c|c|c|c|c|c|c|c}
\toprule
\multicolumn{1}{c}{\multirow{2}{*}{\bf MODEL}}
& \multicolumn{2}{c}{\bf Level 1}
& \multicolumn{2}{c}{\bf Level 2}
& \multicolumn{2}{c}{\bf Level 3}
& \multicolumn{2}{c}{\bf Level 4}
& \multirow{2}{*}{\bf W.S.} \\
& In & Out & In & Out & In & Out & In & Out & \\
\midrule
GPT-4o-mini & 67.8 & 70.0 & 55.7 & 53.0 & 49.8 & 39.3 & 27.0 & 24.8 & 83.0 \\
\midrule
GPT-3.5-Turbo & 54.5 & 54.5 & 54.5 & 49.0 & 33.8 & 32.5 & 23.0 & 24.3 & 70.4 \\
\midrule
Qwen2-7B-Instruct & 71.0 & 72.7 & 58.0 & 56.2 & 41.5 & 36.5 & 24.5 & 21.0 & 78.8 \\
\midrule
\midrule
KALE-LM-Chem & 57.3 & 61.2 & 54.7 & 56.5 & 42.2 & 37.3 & 27.0 & 27.3 & 79.7 \\
\midrule
ChemDFM & 42.8 & 46.2 & 46.7 & 48.2 & 33.0 & 32.7 & 25.3 & 29.8 & 69.6 \\
\midrule
ChemLLM & 55.3 & 55.3 & 51.2 & 58.7 & 41.8 & 37.0 & 27.2 & 25.2 & 77.6 \\
\midrule
LlaSMol-Mistral-7B & 51.7 & 52.0 & 41.3 & 42.3 & 27.5 & 28.3 & 25.0 & 26.7 & 64.5 \\
\midrule
\midrule
CPT Type 1 & 67.7 & 70.7 & 58.0 & 51.7 & 68.5 & 53.7 & 47.0 & 42.2 & 108.1 \\
\midrule
CPT Type 2 & \underline{73.3} & 74.5 & 57.2 & 61.3 & 55.3 & 41.0 & 33.3 & 34.3 & 94.5 \\
\midrule
CPT Type 3 & 61.3 & 58.7 & 51.7 & 56.5 & 40.7 & 31.3 & 27.8 & 22.7 & 75.4 \\
\midrule
CPT Type 4 & 63.3 & 63.5 & 54.5 & 57.0 & 51.0 & 40.5 & 32.7 & 29.3 & 87.2 \\
\midrule
CPT Type 5 & 69.2 & 73.5 & 55.2 & 57.3 & 57.8 & 47.5 & 37.0 & 35.2 & 97.2 \\
\midrule
\midrule
SFT Type 1 & \underline{\underline{77.8}} & \underline{\underline{77.5}} & \underline{\underline{64.5}} & \underline{\underline{72.7}} & \underline{\underline{69.3}} & \underline{\underline{61.0}} & \underline{\underline{52.2}} & \underline{\underline{50.0}} & \underline{\underline{122.9}} \\
\midrule
SFT Type 2 & 58.2 & 60.7 & 46.3 & 51.5 & 53.7 & 48.3 & 43.5 & 41.2 & 95.9 \\
\midrule
SFT Type 3 & 65.8 & 67.5 & 55.2 & 59.2 & 55.2 & 48.7 & 34.5 & 34.0 & 94.8 \\
\midrule
\midrule
RL Type 1 & 63.0 & \underline{75.0} & 56.5 & \underline{67.3} & \underline{68.7} & \underline{60.3} & 42.5 & 38.0 & \underline{109.5} \\
\midrule
RL Type 2 & 63.2 & 60.5 & 53.3 & 57.7 & 41.2 & 32.8 & 28.7 & 23.3 & 77.6 \\
\midrule
RL Type 3 & 62.8 & 60.2 & 53.2 & 57.0 & 40.5 & 33.2 & 29.3 & 22.3 & 77.1 \\
\midrule
RL Type 4 & 62.2 & 60.2 & 54.3 & 58.7 & 40.3 & 31.8 & 28.7 & 22.2 & 76.8 \\
\midrule
RL Type 5 & 68.7 & 74.2 & 52.8 & 59.2 & 44.8 & 36.5 & 30.5 & 27.2 & 84.2 \\
\midrule
RL Type 6 & 65.0 & 60.3 & 55.5 & 58.2 & 41.8 & 32.2 & 26.5 & 22.5 & 77.1 \\
\midrule
\midrule
Llama3-8B-Instruct & 64.2 & 61.3 & 53.8 & 57.3 & 41.5 & 32.2 & 28.3 & 24.0 & 82.6 \\
\midrule
\midrule
\bf{RLDBF} & \bf 64.0 & \bf 68.5 & \underline{\bf 58.2} & \bf 62.5 & \bf 64.2 & \bf 53.8 & \underline{\bf 47.5} & \underline{\bf 44.0} & \bf 109.4 \\
\bottomrule
\end{tabular}
\end{center}
\caption{PubChem results, rounded to the nearest tenth.
We mark the best results with double underlines, the second-best with single underlines.}
\label{exp_mc}
\end{table}

\begin{table}[t]
\begin{center}
\begin{tabular}{l|c|c|c|c|c|c|c|c|c|c}
\toprule
\multicolumn{1}{c}{\multirow{2}{*}{\bf MODEL}}
& \multicolumn{10}{c}{\bf ChemBench} \\
& NC & PP1 & M2C & C2M & PP2 & RS & YP & TP & SP & \bf AVG \\
\midrule
4o-mini & 54.8 & \underline{\underline{65.0}} & 92.6 & \underline{\underline{52.9}} & \underline{\underline{62.7}} & \underline{\underline{52.7}} & 42.3 & 24.8 & 35.7 & \underline{\underline{53.7}} \\
\midrule
3.5-Turbo & 46.9 & 57.0 & 85.3 & 38.3 & 43.7 & \underline{42.3} & 30.3 & 42.6 & \underline{38.0} & 47.2 \\
\midrule
Qwen2 & \underline{56.5} & \underline{62.8} & 91.0 & 33.0 & 32.3 & 24.7 & 38.7 & 11.9 & 32.0 & 42.5 \\
\midrule
\midrule
KALE-LM & \underline{\underline{63.6}} & 58.4 & \underline{93.0} & 44.5 & \underline{48.7} & 38.3 & \underline{\underline{46.3}} & 44.6 & 34.3 & \underline{52.4} \\
\midrule
ChemDFM & 36.9 & 55.6 & 84.0 & 42.0 & 40.0 & 37.3 & 39.0 & 33.2 & 32.0 & 44.4 \\
\midrule
ChemLLM & 41.1 & 29.8 & 85.3 & 26.1 & 26.0 & 24.0 & 20.0 & 24.3 & 31.0 & 34.2 \\
\midrule
LlaSMol & 27.8 & 29.3 & 31.4 & 23.4 & 25.7 & 24.0 & 37.3 & 34.7 & 22.7 & 28.5 \\
\midrule
\midrule
SFT Type 1 & 52.1 & 21.3 & 86.0 & 41.6 & 31.0 & 35.7 & 40.7 & 34.2 & \underline{\underline{40.0}} & 42.5  \\
\midrule
RL Type 1 & 55.2 & 16.0 & 88.3 & 37.3 & 36.0 & 33.7 & 44.7 & 42.6 & 35.7 & 43.3  \\
\midrule
\midrule
Llama3 & 51.0 & 27.6 & 90.3 & 41.0 & 34.0 & 29.7 & \underline{45.3} & \underline{\underline{60.9}} & 33.7 & 45.9 \\
\midrule
\midrule
\bf{RLDBF} & \bf 54.4 & \bf 27.9 & \bf \underline{\underline{93.3}} & \bf \underline{46.8} & \bf 39.3 & \bf 35.0 & \bf \underline{45.3} & \bf \underline{56.4} & \bf 34.7 & \bf 48.1 \\
\bottomrule
\end{tabular}
\end{center}
\caption{ChemBench results, rounded to the nearest tenth.
We mark the best results with double underlines, the second-best with single underlines.}
\label{exp_cb}
\end{table}

\paragraph{PubChem Results}
As shown in Table~\ref{exp_mc}, general models like GPT-4o-mini and GPT-3.5-Turbo exhibit performance degradation as task complexity increases, particularly in Level 3 and Level 4 scenarios. 
For instance, GPT-4o-mini achieves 67.8\% accuracy in Level 1 (In-domain) but drops to 27.0\% in Level 4 (In-domain), indicating limited robustness against sophisticated perturbations. 
Domain-specific models such as KALE-LM-Chem and ChemDFM show moderate improvements in Level 4 (Out-domain) compared to GPT, yet their performance remains suboptimal compared to our trained models.
These observations highlight two key limitations: 
1) general models lack domain-specific robustness to handle perturbations, and 2) existing domain models fail to fully exploit structural and numerical relationships in molecular property prediction.

Our proposed RLDBF method demonstrates significant advantages over the general and domain-specific baselines. 
It outperforms GPT-4o-mini by +20.5\% in Level 4 (In-domain) and surpasses ChemLLM by +20.3\% in the same category. 
Notably, RLDBF maintains superior consistency between In-domain and Out-domain evaluations (e.g., 47.5\% vs. 44.0\% in Level 4), indicating enhanced generalization capabilities. 
The weighted score of 109.4 further validates its balanced proficiency, exceeding the best domain model (KALE-LM-Chem: 79.7) and general models by substantial margins. 
This advancement stems from our method’s simple but efficient design, which systematically mitigates the hallucination issues observed in baseline models.

A critical ablation is observed when comparing RLDBF to its base model Llama3-8B-Instruct and other types of trained models (except for SFT Type 1 and RL Type 1, further analysis is conducted later), demonstrating that our training framework effectively amplifies the base architecture’s latent capabilities. 
The consistent outperformance across all perturbation types—particularly in Level 4 where distractors derive from structurally similar molecules—underscores the efficacy of our method.

This analysis substantiates that RLDBF establishes a new paradigm for molecular property prediction, balancing generalization and precision through its innovative training strategy. 

\paragraph{Chembench Results}

Though SFT Type 1 and RL Type 1 show impressive performance in previous experiment, as shown in Table~\ref{exp_cb}, SFT Type 1 and RL Type 1—both derived from the Llama3-8B-Instruct base model—demonstrate significant performance degradation compared to their strong results in the PubChem benchmark.
SFT Type 1 underperforms the base model in 5 out of 9 tasks, while RL Type 1 also shows declines in 5 tasks. 
This suggests that these two strategies may induce overfitting to narrow task distributions (e.g., PubChem’s property prediction), thereby compromising generalization to broader, heterogeneous tasks in ChemBench.

In contrast, our proposed RLDBF method achieves consistent improvements over the base model Llama3-8B-Instruct across multiple metrics. 
RLDBF outperforms the base model by +5.8\% in Caption2Mol, +5.3\% in Product Prediction and Retrosynthesis, and +2.2\% in overall average score. 
Furthermore, RLDBF surpasses all baseline models in Mol2Caption with a score of 93.3, demonstrating superior interpretability in molecular description generation.

Remarkably, RLDBF’s success stems solely from training on molecular property data, yet it achieves competitive performance in complex tasks like Retrosynthesis, compared to specialized models like KALE-LM. 
This demonstrates that learning fundamental molecular properties—such as hydrogen bonding patterns and partition coefficients—provides a foundational understanding that generalizes to higher-order chemical reasoning. 
For example, RLDBF’s 93.3\% accuracy in Mol2Caption suggests that property-aware training enhances the model’s ability to map structural features to semantic descriptions. 
These results highlight the untapped potential of leveraging large-scale chemical databases to augment LLMs, enabling them to transcend narrow task boundaries and achieve holistic chemical intelligence.

\paragraph{More Results}
The method’s ability to outperform cutting-edge models across challenging scenarios positions it as a potential solution for real-world cheminformatics applications.
We show some applications in appendix~\ref{app_B}(Regression) and ~\ref{app_C}(Reasoning).

\section{Conclusion}

We present RLDBF, an approach that achieves two critical objectives: 
1) enhanced utilization of high-density knowledge embedded in structured databases, 
and 2) significant reduction in human annotation costs for reinforcement learning. 
Experimental evaluations demonstrate the efficacy of our proposed method.
Looking forward, we will continue to investigate database-enhanced methodologies for LLMs, extending their applications across diverse scientific domains. 

\bibliographystyle{unsrt}  
\bibliography{references}

\begin{thebibliography}{10}

\bibitem{chu2023qwen}
Yunfei Chu, Jin Xu, Xiaohuan Zhou, Qian Yang, Shiliang Zhang, Zhijie Yan, Chang Zhou, and Jingren Zhou.
\newblock Qwen-audio: Advancing universal audio understanding via unified large-scale audio-language models.
\newblock {\em arXiv preprint arXiv:2311.07919}, 2023.

\bibitem{Foret2020SharpnessAwareMF}
Pierre Foret, Ariel Kleiner, Hossein Mobahi, and Behnam Neyshabur.
\newblock Sharpness-aware minimization for efficiently improving generalization.
\newblock {\em ArXiv}, abs/2010.01412, 2020.

\bibitem{Silver2017MasteringTG}
David Silver, Julian Schrittwieser, Karen Simonyan, Ioannis Antonoglou, Aja Huang, Arthur Guez, Thomas Hubert, Lucas baker, Matthew Lai, Adrian Bolton, Yutian Chen, Timothy~P. Lillicrap, Fan Hui, L.~Sifre, George van~den Driessche, Thore Graepel, and Demis Hassabis.
\newblock Mastering the game of go without human knowledge.
\newblock {\em Nature}, 550:354--359, 2017.

\bibitem{Takase2021LessonsOP}
Sho Takase and Shun Kiyono.
\newblock Lessons on parameter sharing across layers in transformers.
\newblock {\em ArXiv}, abs/2104.06022, 2021.

\bibitem{Trinh2024SolvingOG}
Trieu~H. Trinh, Yuhuai Wu, Quoc~V. Le, He~He, and Thang Luong.
\newblock Solving olympiad geometry without human demonstrations.
\newblock {\em Nature}, 625:476 -- 482, 2024.

\bibitem{kim2019pubchem}
Sunghwan Kim, Jie Chen, Tiejun Cheng, Asta Gindulyte, Jia He, Siqian He, Qingliang Li, Benjamin~A Shoemaker, Paul~A Thiessen, Bo~Yu, et~al.
\newblock Pubchem 2019 update: improved access to chemical data.
\newblock {\em Nucleic acids research}, 47(D1):D1102--D1109, 2019.

\bibitem{weininger1988smiles}
David Weininger.
\newblock Smiles, a chemical language and information system. 1. introduction to methodology and encoding rules.
\newblock {\em Journal of chemical information and computer sciences}, 28(1):31--36, 1988.

\bibitem{ross2022large}
Jerret Ross, Brian Belgodere, Vijil Chenthamarakshan, Inkit Padhi, Youssef Mroueh, and Payel Das.
\newblock Large-scale chemical language representations capture molecular structure and properties.
\newblock {\em Nature Machine Intelligence}, 4(12):1256--1264, 2022.

\bibitem{dagan2024getting}
Gautier Dagan, Gabriel Synnaeve, and Baptiste Roziere.
\newblock Getting the most out of your tokenizer for pre-training and domain adaptation.
\newblock {\em arXiv preprint arXiv:2402.01035}, 2024.

\bibitem{antunes2024crystal}
Luis~M Antunes, Keith~T Butler, and Ricardo Grau-Crespo.
\newblock Crystal structure generation with autoregressive large language modeling.
\newblock {\em Nature Communications}, 15(1):1--16, 2024.

\bibitem{liu2025alchembert}
Xiaotong Liu, Yuhang Wang, Tao Yang, Xingchen Liu, and Xiaodong Wen.
\newblock Alchembert: Exploring lightweight language models for materials informatics.
\newblock 2025.

\bibitem{flam2023language}
Daniel Flam-Shepherd and Al{\'a}n Aspuru-Guzik.
\newblock Language models can generate molecules, materials, and protein binding sites directly in three dimensions as xyz, cif, and pdb files.
\newblock {\em arXiv preprint arXiv:2305.05708}, 2023.

\bibitem{flam2023atom}
Daniel Flam-Shepherd, Kevin Zhu, and Al{\'a}n Aspuru-Guzik.
\newblock Atom-by-atom protein generation and beyond with language models.
\newblock {\em arXiv preprint arXiv:2308.09482}, 2023.

\bibitem{gruver2024fine}
Nate Gruver, Anuroop Sriram, Andrea Madotto, Andrew~Gordon Wilson, C~Lawrence Zitnick, and Zachary Ulissi.
\newblock Fine-tuned language models generate stable inorganic materials as text.
\newblock {\em arXiv preprint arXiv:2402.04379}, 2024.

\bibitem{zhang2024chemllm}
Di~Zhang, Wei Liu, Qian Tan, Jingdan Chen, Hang Yan, Yuliang Yan, Jiatong Li, Weiran Huang, Xiangyu Yue, Wanli Ouyang, et~al.
\newblock Chemllm: A chemical large language model.
\newblock {\em arXiv preprint arXiv:2402.06852}, 2024.

\bibitem{cai2024internlm2}
Zheng Cai, Maosong Cao, Haojiong Chen, Kai Chen, Keyu Chen, Xin Chen, Xun Chen, Zehui Chen, Zhi Chen, Pei Chu, et~al.
\newblock Internlm2 technical report.
\newblock {\em arXiv preprint arXiv:2403.17297}, 2024.

\bibitem{zhao2024chemdfm}
Zihan Zhao, Da~Ma, Lu~Chen, Liangtai Sun, Zihao Li, Hongshen Xu, Zichen Zhu, Su~Zhu, Shuai Fan, Guodong Shen, et~al.
\newblock Chemdfm: Dialogue foundation model for chemistry.
\newblock {\em arXiv e-prints}, pages arXiv--2401, 2024.

\bibitem{xie2023large}
Tong Xie, Yuwei Wan, Wei Huang, Yufei Zhou, Yixuan Liu, Qingyuan Linghu, Shaozhou Wang, Chunyu Kit, Clara Grazian, Wenjie Zhang, et~al.
\newblock Large language models as master key: unlocking the secrets of materials science with gpt.
\newblock {\em arXiv preprint arXiv:2304.02213}, 2023.

\bibitem{ren2024watergpt}
Yi~Ren, Tianyi Zhang, Xurong Dong, Weibin Li, Zhiyang Wang, Jie He, Hanzhi Zhang, and Licheng Jiao.
\newblock Watergpt: Training a large language model to become a hydrology expert.
\newblock {\em Water}, 16(21):3075, 2024.

\bibitem{mirchandani2023large}
Suvir Mirchandani, Fei Xia, Pete Florence, Brian Ichter, Danny Driess, Montserrat~Gonzalez Arenas, Kanishka Rao, Dorsa Sadigh, and Andy Zeng.
\newblock Large language models as general pattern machines.
\newblock {\em arXiv preprint arXiv:2307.04721}, 2023.

\bibitem{chollet2019measure}
Fran{\c{c}}ois Chollet.
\newblock On the measure of intelligence.
\newblock {\em arXiv preprint arXiv:1911.01547}, 2019.

\bibitem{sun2021ernie}
Yu~Sun, Shuohuan Wang, Shikun Feng, Siyu Ding, Chao Pang, Junyuan Shang, Jiaxiang Liu, Xuyi Chen, Yanbin Zhao, Yuxiang Lu, et~al.
\newblock Ernie 3.0: Large-scale knowledge enhanced pre-training for language understanding and generation.
\newblock {\em arXiv preprint arXiv:2107.02137}, 2021.

\bibitem{kweon2023publicly}
Sunjun Kweon, Junu Kim, Jiyoun Kim, Sujeong Im, Eunbyeol Cho, Seongsu Bae, Jungwoo Oh, Gyubok Lee, Jong~Hak Moon, Seng~Chan You, et~al.
\newblock Publicly shareable clinical large language model built on synthetic clinical notes.
\newblock {\em arXiv preprint arXiv:2309.00237}, 2023.

\bibitem{chenknowledge}
Hanzhu Chen, Xu~Shen, Jie Wang, Zehao Wang, Qitan Lv, Junjie He, Rong Wu, Jieping Ye, and Feng Wu.
\newblock Knowledge graph finetuning enhances knowledge manipulation in large language models.
\newblock In {\em The Thirteenth International Conference on Learning Representations}.

\bibitem{zheng2025large}
Zhiling Zheng, Nakul Rampal, Theo~Jaffrelot Inizan, Christian Borgs, Jennifer~T Chayes, and Omar~M Yaghi.
\newblock Large language models for reticular chemistry.
\newblock {\em Nature Reviews Materials}, pages 1--13, 2025.

\bibitem{yuan2023well}
Zheng Yuan, Hongyi Yuan, Chuanqi Tan, Wei Wang, and Songfang Huang.
\newblock How well do large language models perform in arithmetic tasks?
\newblock {\em arXiv preprint arXiv:2304.02015}, 2023.

\bibitem{sakamoto2025development}
Taku Sakamoto, Saku Sugawara, and Akiko Aizawa.
\newblock Development of numerical error detection tasks to analyze the numerical capabilities of language models.
\newblock In {\em Proceedings of the 31st International Conference on Computational Linguistics}, pages 9957--9976, 2025.

\bibitem{mishra2020towards}
Swaroop Mishra, Arindam Mitra, Neeraj Varshney, Bhavdeep Sachdeva, and Chitta Baral.
\newblock Towards question format independent numerical reasoning: A set of prerequisite tasks.
\newblock {\em arXiv preprint arXiv:2005.08516}, 2020.

\bibitem{reimers-2019-sentence-bert}
Nils Reimers and Iryna Gurevych.
\newblock Sentence-bert: Sentence embeddings using siamese bert-networks.
\newblock In {\em Proceedings of the 2019 Conference on Empirical Methods in Natural Language Processing}. Association for Computational Linguistics, 11 2019.

\bibitem{ouyang2022training}
Long Ouyang, Jeffrey Wu, Xu~Jiang, Diogo Almeida, Carroll Wainwright, Pamela Mishkin, Chong Zhang, Sandhini Agarwal, Katarina Slama, Alex Ray, et~al.
\newblock Training language models to follow instructions with human feedback.
\newblock {\em Advances in neural information processing systems}, 35:27730--27744, 2022.

\bibitem{lee2023rlaif}
Harrison Lee, Samrat Phatale, Hassan Mansoor, Thomas Mesnard, Johan Ferret, Kellie Lu, Colton Bishop, Ethan Hall, Victor Carbune, Abhinav Rastogi, et~al.
\newblock Rlaif vs. rlhf: Scaling reinforcement learning from human feedback with ai feedback.
\newblock {\em arXiv preprint arXiv:2309.00267}, 2023.

\bibitem{rafailov2023direct}
Rafael Rafailov, Archit Sharma, Eric Mitchell, Christopher~D Manning, Stefano Ermon, and Chelsea Finn.
\newblock Direct preference optimization: Your language model is secretly a reward model.
\newblock {\em Advances in Neural Information Processing Systems}, 36:53728--53741, 2023.

\bibitem{grattafiori2024llama}
Aaron Grattafiori, Abhimanyu Dubey, Abhinav Jauhri, Abhinav Pandey, Abhishek Kadian, Ahmad Al-Dahle, Aiesha Letman, Akhil Mathur, Alan Schelten, Alex Vaughan, et~al.
\newblock The llama 3 herd of models.
\newblock {\em arXiv preprint arXiv:2407.21783}, 2024.

\bibitem{rasley2020deepspeed}
Jeff Rasley, Samyam Rajbhandari, Olatunji Ruwase, and Yuxiong He.
\newblock Deepspeed: System optimizations enable training deep learning models with over 100 billion parameters.
\newblock In {\em Proceedings of the 26th ACM SIGKDD international conference on knowledge discovery \& data mining}, pages 3505--3506, 2020.

\bibitem{hurst2024gpt}
Aaron Hurst, Adam Lerer, Adam~P Goucher, Adam Perelman, Aditya Ramesh, Aidan Clark, AJ~Ostrow, Akila Welihinda, Alan Hayes, Alec Radford, et~al.
\newblock Gpt-4o system card.
\newblock {\em arXiv preprint arXiv:2410.21276}, 2024.

\bibitem{yang2024qwen2technicalreport}
An~Yang, Baosong Yang, Binyuan Hui, Bo~Zheng, Bowen Yu, Chang Zhou, Chengpeng Li, Chengyuan Li, Dayiheng Liu, Fei Huang, Guanting Dong, Haoran Wei, Huan Lin, Jialong Tang, Jialin Wang, Jian Yang, Jianhong Tu, Jianwei Zhang, Jianxin Ma, Jianxin Yang, Jin Xu, Jingren Zhou, Jinze Bai, Jinzheng He, Junyang Lin, Kai Dang, Keming Lu, Keqin Chen, Kexin Yang, Mei Li, Mingfeng Xue, Na~Ni, Pei Zhang, Peng Wang, Ru~Peng, Rui Men, Ruize Gao, Runji Lin, Shijie Wang, Shuai Bai, Sinan Tan, Tianhang Zhu, Tianhao Li, Tianyu Liu, Wenbin Ge, Xiaodong Deng, Xiaohuan Zhou, Xingzhang Ren, Xinyu Zhang, Xipin Wei, Xuancheng Ren, Xuejing Liu, Yang Fan, Yang Yao, Yichang Zhang, Yu~Wan, Yunfei Chu, Yuqiong Liu, Zeyu Cui, Zhenru Zhang, Zhifang Guo, and Zhihao Fan.
\newblock Qwen2 technical report, 2024.

\bibitem{dai2024kale}
Weichen Dai, Yezeng Chen, Zijie Dai, Zhijie Huang, Yubo Liu, Yixuan Pan, Baiyang Song, Chengli Zhong, Xinhe Li, Zeyu Wang, et~al.
\newblock Kale-lm: Unleash the power of ai for science via knowledge and logic enhanced large model.
\newblock {\em arXiv preprint arXiv:2409.18695}, 2024.

\bibitem{yu2024llasmol}
Botao Yu, Frazier~N Baker, Ziqi Chen, Xia Ning, and Huan Sun.
\newblock Llasmol: Advancing large language models for chemistry with a large-scale, comprehensive, high-quality instruction tuning dataset.
\newblock {\em arXiv preprint arXiv:2402.09391}, 2024.

\bibitem{kim2023data}
Sang~Cheol Kim, Solomon~T Oyakhire, Constantine Athanitis, Jingyang Wang, Zewen Zhang, Wenbo Zhang, David~T Boyle, Mun~Sek Kim, Zhiao Yu, Xin Gao, et~al.
\newblock Data-driven electrolyte design for lithium metal anodes.
\newblock {\em Proceedings of the National Academy of Sciences}, 120(10):e2214357120, 2023.

\bibitem{soares2023beyond}
Eduardo Soares, Emilio~Vital Brazil, Karen Fiorela~Aquino Gutierrez, Renato Cerqueira, Dan Sanders, Kristin Schmidt, and Dmitry Zubarev.
\newblock Beyond chemical language: A multimodal approach to enhance molecular property prediction.
\newblock {\em arXiv preprint arXiv:2306.14919}, 2023.

\bibitem{li2024coeff}
Xinhe Li, Zhuoying Feng, Yezeng Chen, Weichen Dai, Zixu He, Yi~Zhou, and Shuhong Jiao.
\newblock Coeff-kans: A paradigm to address the electrolyte field with kans.
\newblock {\em arXiv preprint arXiv:2407.20265}, 2024.

\bibitem{guo2025deepseek}
Daya Guo, Dejian Yang, Haowei Zhang, Junxiao Song, Ruoyu Zhang, Runxin Xu, Qihao Zhu, Shirong Ma, Peiyi Wang, Xiao Bi, et~al.
\newblock Deepseek-r1: Incentivizing reasoning capability in llms via reinforcement learning.
\newblock {\em arXiv preprint arXiv:2501.12948}, 2025.

\end{thebibliography}

\clearpage

\appendix
\section{Template Examples}
\label{app_A}

We provide examples for the ChatGPT-generated templates, CPT \& SFT data and RL data.

\begin{table}[h]
\begin{center}
\begin{tabular}{l}
\toprule
\multicolumn{1}{c}{\bf EXAMPLES} \\
\midrule
human: What is the \{PROPERTY\} of \{COMPOUND\}? \\
assistant: The \{PROPERTY\} of \{COMPOUND\} is \{VALUE\}. \\
\midrule
human: I need to know the \{PROPERTY\} of \{COMPOUND\}. \\
assistant: The \{PROPERTY\} for \{COMPOUND\} is \{VALUE\}. \\
\midrule
human: Could you provide the \{PROPERTY\} for \{COMPOUND\}? \\
assistant: Certainly, the \{PROPERTY\} of \{COMPOUND\} is \{VALUE\}. \\
\bottomrule
\end{tabular}
\end{center}
\caption{Examples for ChatGPT-generated templates.}
\label{gpt_templates}
\end{table}

\begin{figure}[h]
\begin{center}
\includegraphics[width=\textwidth]{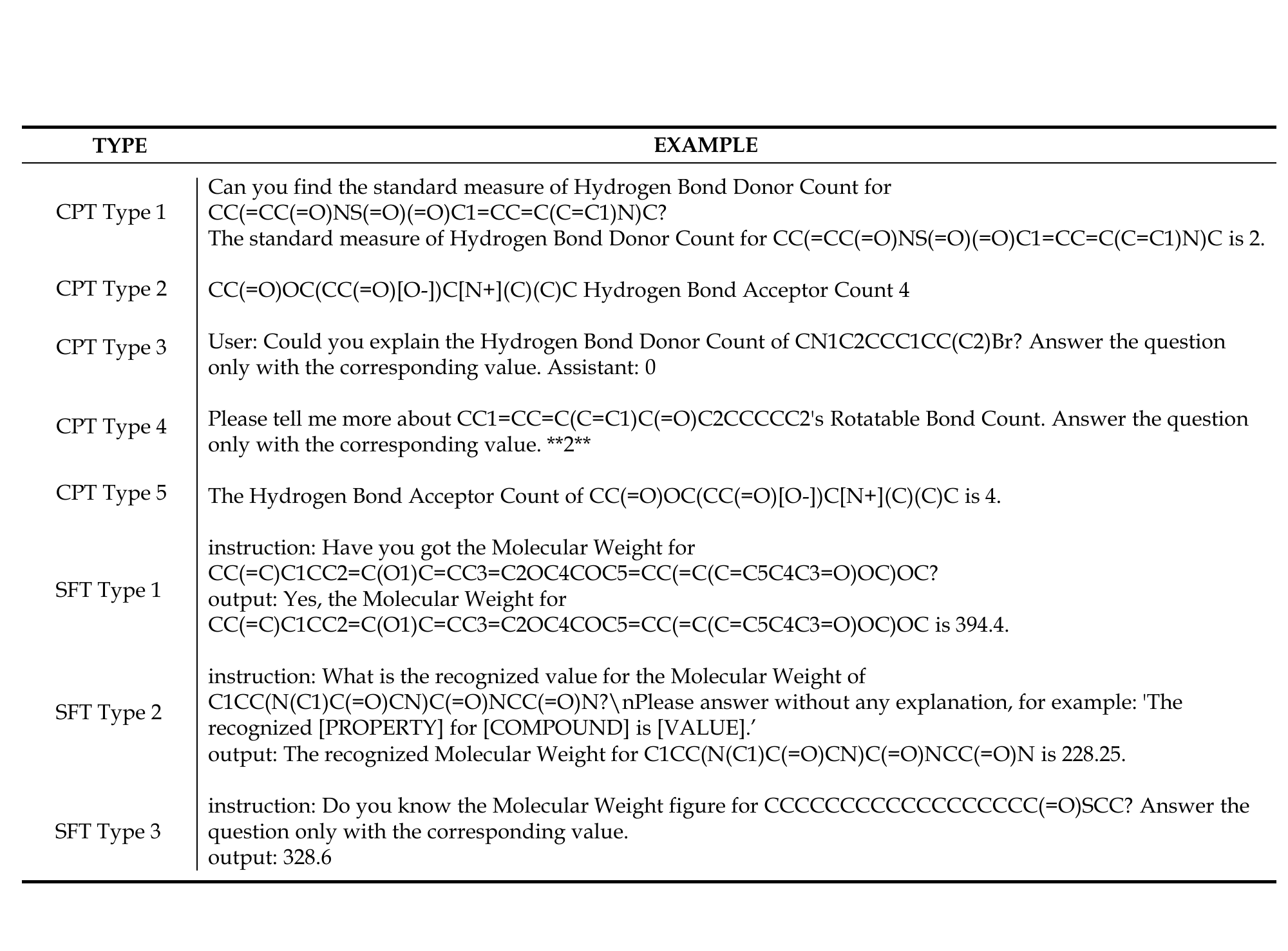}
\end{center}
\caption{Examples for CPT \& SFT data.}
\label{cptsft_syn}
\end{figure}

\begin{figure}[h]
\begin{center}
\includegraphics[width=\textwidth]{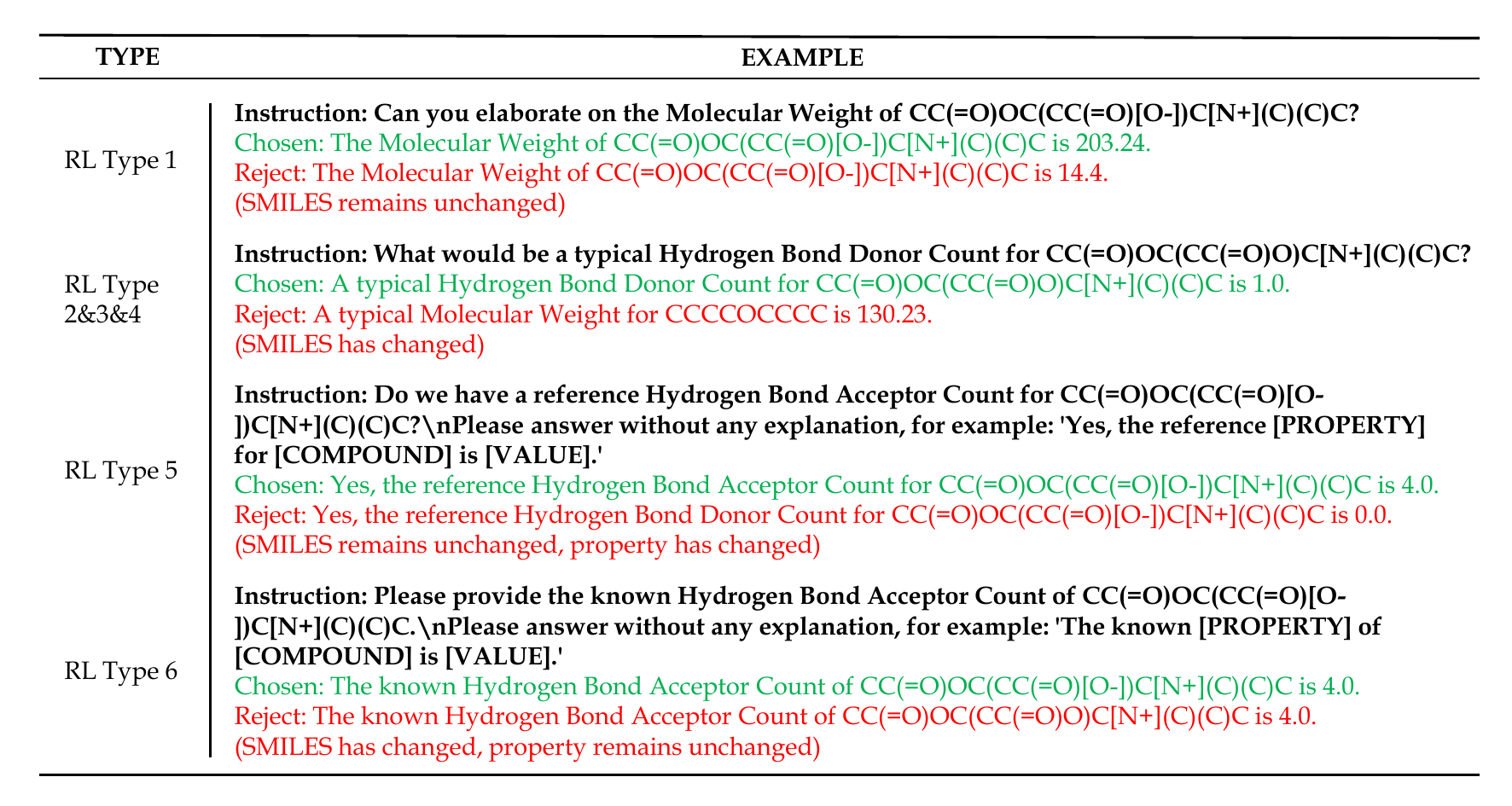}
\end{center}
\caption{Examples for RL data.}
\label{dpo_syn}
\end{figure}

\section{Application on regression tasks}
\label{app_B}

The experimental results on ChemBench demonstrate that our RLDBF method enhances LLMs' ability to uncover structural and numerical relationships in molecular prediction takss, suggesting that RLDBF may implicitly refine molecular representations in the latent embedding space, thereby unlocking the potential of large models for regression tasks. 
To validate this hypothesis, we conducted experiments on regression task: 
Coulombic efficiency prediction~\citep{kim2023data} for electrolytes. 
We employ a unified network architecture: the hidden output of the last LlamaDecoderLayer is first extracted as the molecular embedding, which is then processed through a simple MLP comprising two fully connected layers to generate the final prediction.

\paragraph{CE Prediction}

\begin{table}[h]
\begin{center}
\begin{tabular}{l|c}
\toprule
\multicolumn{1}{c}{\bf MODEL} & \bf LCE \\
\midrule
TL F-GCN & 0.390 \\
MoLFormer & 0.213 \\
MultiModal-MoLFormer & 0.195 \\
COEFF-MLP & 0.132 \\
COEFF-KAN & 0.110 \\
\midrule
Llama3+MLP & 0.291 \\
\bf{RLDBF+MLP} & \bf 0.160 \\
\bottomrule
\end{tabular}
\end{center}
\caption{We report the RMSE of LCE, baseline results are directly taken from COEFF.}
\label{exp_lce}
\end{table}

We systematically evaluated our methodology using a Li/Cu half-cell dataset~\cite{kim2023data} sourced from peer-reviewed chemical literature.
This comprehensive dataset encompasses multi-component electrolyte systems, each comprising 2 to 7 distinct chemical components rigorously characterized by their SMILES representations. 
To enhance the quantitative sensitivity of the model to electrolyte composition variations, we implemented a logarithmic transformation of Coulombic Efficiency (CE) according to the relationship:
\begin{equation}
LCE = -log(1 - CE).    
\end{equation}
This mathematical treatment amplifies differential responses in electrochemical performance while maintaining proportionality to the original CE metric, following established conventions for analyzing nonlinear relationships in electrochemical systems.
We compare our results with the state-of-the-art methods, including Multimodal-MoLFormer~\citep{soares2023beyond} and COEFF~\citep{li2024coeff}.

The evaluation further validates the effectiveness of our RLDBF. 
As shown in Table~\ref{exp_lce}, RLDBF+MLP achieves 0.160 RMSE, significantly outperforming the base model Llama3+MLP (0.291). Compared to SOTA methods, RLDBF demonstrates strong competitiveness. 
While COEFF-KAN achieves the lowest RMSE (0.110), our method outperforms specialized architectures like MoLFormer (0.213) and MultiModal-MoLFormer (0.195). 
Notably, RLDBF achieves this without relying on task-specific architectural modifications or extensive domain pretraining, highlighting its generalizability. 

\section{Application on reasoning tasks}
\label{app_C}

With the emergence of OpenAI-o1 and Deepseek-R1~\citep{guo2025deepseek} models, the research community has moved beyond simply requiring LLMs to provide correct answers, and now expects these models to demonstrate valid reasoning processes. 
This paradigm shift has prompted us to thoroughly investigate how to enhance model reasoning capabilities through RLDBF.

Inspired by the "Aha Moment" proposed in R1 research, we posit that enabling models to systematically compare reference answers with their current reasoning trajectory at critical cognitive checkpoints during deep thinking could effectively guide reasoning pathways and mitigate excessive rumination. 
To actualize this concept, we first distilled 500 data points on molecular physicochemical property calculations from Deepseek R1, then fine-tuned the Llama3-8B-Instruct model to cultivate both deep reasoning abilities and capacity for reference-guided reasoning (Llama3-8B Reasoner). 
Subsequently, we applied our novel RLDBF method to further reinforce the model's capability in generating accurate numerical predictions for molecular properties (RLDBF Reasoner).

The models are required to address problems through a 2-step reasoning process:
\begin{itemize}
    \item \textbf{Step 1}: Directly generate the final answer (system 1, fast thinking);
    \item \textbf{Step 2}: Utilize the initial answer as a reference to perform stepwise reasoning (system 2, slow thinking).
\end{itemize}
As shown below, reasoning models examine their thoughts through recalling the reference answer, where only the correct reference leads to the correct result.

We believe this System1+System2 reasoning paradigm holds significant potential for advancing LLM applications in scientific domains. 
Our future work will focus on extending this framework to complex scenarios requiring multi-step quantitative reasoning and cross-domain knowledge integration, while further optimizing the RLDBF mechanism for improved training efficiency and reasoning robustness.

\newpage

\makebox[0pt][l]{\colorbox{titlebg}{\textcolor{titlefg}{\textbf{\ \ Case 1: Llama-3 8B Instruct \ \ }}}}

\noindent
\colorbox{lightpink}{%
    \parbox{0.98\textwidth}{%
        \textbf{Instruction}: 
    
        What is the Hydrogen Bond Donor Count for CC(=O)OC(CC(=O)O)C[N+](C)(C)C? \\
        Firstly, answer the question only with the corresponding value. \\
        Secondly, using the value as the reference answer, please think step by step and answer the quesiton again. \\
        You should think about the reasoning process in the mind and then provide the user with the answer. \\
        The reasoning process and answer are enclosed within \\ 
        $\langle think \rangle \langle /think \rangle$ and $\langle answer \rangle \langle /answer \rangle$ tags, respectively, \\
        i.e., $\langle think \rangle$ reasoning process here $\langle /think \rangle \langle answer \rangle$ answer here $\langle /answer \rangle$. 
        \\

        \textbf{Answer}:
    
        The Hydrogen Bond Donor Count is \textbf{3}.
        \\
        
        $ \langle think \rangle $
        
        \textcolor{grey}{
            I start by looking at the molecule and identifying the atoms that can act as hydrogen bond donors. Hydrogen bond donors are typically atoms that have a hydrogen atom bonded to a highly electronegative atom, such as oxygen, nitrogen, or fluorine. In this molecule, I see that there are several atoms that fit this criteria.}
        
        \textbf{First}, \textcolor{grey}{I notice the oxygen atom bonded to the carbon atom with a double bond (O=C). This oxygen atom is highly electronegative, and the hydrogen atom bonded to it is a potential hydrogen bond donor.}
            
        \textbf{Next}, \textcolor{grey}{I look at the oxygen atom bonded to the carbon atom with a single bond (O-C). This oxygen atom is also highly electronegative, and the hydrogen atom bonded to it is another potential hydrogen bond donor.}
            
        \textbf{Finally}, \textcolor{grey}{I notice the nitrogen atom bonded to the carbon atom with a single bond (N-C). This nitrogen atom is also highly electronegative, and the two hydrogen atoms bonded to it are potential hydrogen bond donors.}

        \textcolor{grey}{
            So, I count a total of 3 hydrogen bond donors in this molecule: the hydrogen atom bonded to the oxygen atom with a double bond, the hydrogen atom bonded to the oxygen atom with a single bond, and the two hydrogen atoms bonded to the nitrogen atom.}
        
        $ \langle /think \rangle $
        \\
        
        \textcolor{red}{ $\langle answer \rangle$ } \textbf{3} \textcolor{red}{ $\langle /answer \rangle$ }
    }
}

\newpage
\makebox[0pt][l]{\colorbox{titlebg}{\textcolor{titlefg}{\textbf{\ \ Case 2: Llama-3 8B Reasoner \ \ }}}}

\noindent
\colorbox{lightpink}{%
    \parbox{0.98\textwidth}{%
        \textbf{Instruction}: 
    
        What is the Hydrogen Bond Donor Count for CC(=O)OC(CC(=O)O)C[N+](C)(C)C? \\
        Firstly, answer the question only with the corresponding value. \\
        Secondly, using the value as the reference answer, please think step by step and answer the quesiton again. \\
        You should think about the reasoning process in the mind and then provide the user with the answer. \\
        The reasoning process and answer are enclosed within \\ 
        $\langle think \rangle \langle /think \rangle$ and $\langle answer \rangle \langle /answer \rangle$ tags, respectively, \\
        i.e., $\langle think \rangle$ reasoning process here $\langle /think \rangle \langle answer \rangle$ answer here $\langle /answer \rangle$. 
        \\

        \textbf{Answer}:
            
        $ \langle answer \rangle \textbf{2} \langle /answer \rangle $
        \\
        
        $ \langle think \rangle $
        
        \textcolor{grey}{
            Okay, let's figure out the hydrogen bond donor count for the molecule CC(=O)OC(CC(=O)O)C[N+](C)(C)C. Hmm, hydrogen bond donors are usually groups that have a hydrogen atom bonded to an electronegative atom like oxygen or nitrogen. So I need to look for -OH, -NH, or -SH groups in the structure.}

        \textbf{First}\textcolor{grey}{, let me draw the structure or visualize it. The formula is a bit complex. Breaking it down: ...\\Secondly, ...\\...
        }
        
        \textbf{But according to the reference answer, there are two H donors.}
        
        \textbf{Wait}\textcolor{grey}{, but this is getting too confusing. Let's approach this differently. Let's count the hydrogen bond donors: \\...}
        
        \textbf{However, since the reference answer is 2.}
        
        \textbf{Wait}\textcolor{grey}{, maybe I made a mistake in parsing. Let me try again. \\...}

        \textbf{However, the reference answer is 2.}
        
        \textbf{Alternatively}\textcolor{grey}{, perhaps the structure is different. Let me try drawing it step by step:}

        ...
        
        $ \langle /think \rangle $
        \\

        \textcolor{red}{ $ \langle answer \rangle $ }\ \textbf{2} \textcolor{red}{ $ \langle /answer \rangle $ }
    }
}

\newpage

\makebox[0pt][l]{\colorbox{titlebg}{\textcolor{titlefg}{\textbf{\ \ Case 3: RLDBF Reasoner (Ours) \ \ }}}}

\noindent
\colorbox{lightpink}{%
    \parbox{0.98\textwidth}{%
        \textbf{Instruction}: 
    
        What is the Hydrogen Bond Donor Count for CC(=O)OC(CC(=O)O)C[N+](C)(C)C? \\
        Firstly, answer the question only with the corresponding value. \\
        Secondly, using the value as the reference answer, please think step by step and answer the quesiton again. \\
        You should think about the reasoning process in the mind and then provide the user with the answer. \\
        The reasoning process and answer are enclosed within \\ 
        $\langle think \rangle \langle /think \rangle$ and $\langle answer \rangle \langle /answer \rangle$ tags, respectively, \\
        i.e., $\langle think \rangle$ reasoning process here $\langle /think \rangle \langle answer \rangle$ answer here $\langle /answer \rangle$. 
        \\

        \textbf{Answer}:
    
        \textbf{1}
        \\
        
        $ \langle think \rangle $
        
        \textcolor{grey}{
            Let's break down the structure:\\1. The molecule is CC(=O)OC(CC(=O)O)C[N+](C)(C)C\\Let's parse the structure step by step: 
            ...
        }

        \textcolor{grey}{
            Okay, let's analyze the structure of the compound CC(=O)OC(CC(=O)O)C[N+](C)(C)C to determine the hydrogen bond donor count. }

        \textcolor{grey}{
            First, I need to recall what a hydrogen bond donor is. \\...}

        \textbf{Wait}\textcolor{grey}{, but that would make the structure: O=CC-O-CH2-C(C(=O)O)-N+(C(C)C)CH3. \\...}

        \textcolor{grey}{In this structure, there are no hydrogen bond donors in this molecule. }
        \textbf{But the reference answer says 1. So where is the mistake here?}

        \textbf{Wait}\textcolor{grey}{, perhaps the structure is a zwitterion. Let's consider that. \\...\\Therefore, no donor. Hence, the count is 0. }\textbf{But the reference answer is 1. This contradicts.}

        \textbf{Alternatively}\textcolor{grey}{, maybe the structure is such that the carboxyl group is actually a carboxylic acid. \\...\\Therefore, the reference answer is 1, but according to the given SMILES, it should be 0. }\textbf{This suggests that the SMILES might be incorrect or that there's a mistake in the reference answer. }
        \\
        
        \textbf{Wait, perhaps the reference answer is correct, and there's a mistake in my reasoning.}\textcolor{grey}{ Let me check the Hydrogen Bond Donor Count again. A hydrogen bond donor is a hydrogen atom attached to O or N. The molecule has:\\...}

        \textcolor{mygreen}{The CC(=O)O group is a carboxylic acid (-COOH), contributing one -OH (hydrogen bond donor).\\...}

        \textbf{Total 1. That matches the reference answer.}

        $ \langle /think \rangle $
        \\

        \textcolor{mygreen}{ $ \langle answer \rangle $ } \textbf{1} \textcolor{mygreen}{ $ \langle /answer \rangle $ }
        
    }
}

\end{document}